\pdfoutput=1

\documentclass[11pt]{article}

\usepackage[final]{acl}

\usepackage{times}
\usepackage{latexsym}

\usepackage[T1]{fontenc}

\usepackage[utf8]{inputenc}

\usepackage{microtype}

\usepackage{inconsolata}

\usepackage{graphicx}

\newcommand*\methodname{\textsc{LLMBraces}}

\NewDocumentCommand{\ying}{ mO{} }{\textcolor{teal}{\textsuperscript{\textit{Ying}}\textsf{\textbf{\small[#1]}}}}

\NewDocumentCommand{\lifu}{ mO{} }{\textcolor{red}{\textsuperscript{\textit{Lifu}}\textsf{\textbf{\small[#1]}}}}

\usepackage{microtype}
\usepackage{amsmath,amsfonts,bm}

\usepackage{inconsolata}
\usepackage[capitalize]{cleveref}
\crefformat{equation}{Eq.~(#2#1#3)}
\crefname{section}{§}{§§}
\Crefname{section}{§}{§§}
\usepackage{booktabs}       
\usepackage{nicefrac}       
\usepackage{xcolor}         
\usepackage{graphicx}
\usepackage{multirow}
\usepackage{caption}
\usepackage{subcaption}
\usepackage{amssymb}
\usepackage{listings}

\usepackage{colortbl}
\usepackage{array}
\usepackage{environ}

\definecolor{sub-updates}{HTML}{7F6000}
\definecolor{value}{HTML}{41A15E}
\definecolor{key}{HTML}{FB8E8D}
\definecolor{relevance}{HTML}{ED7D31}
\definecolor{sub-updates_bg}{HTML}{FFF2CC}

\crefformat{equation}{Eq.~(#2#1#3)}
\crefname{section}{§}{§§}
\Crefname{section}{§}{§§}

\usepackage{amsmath,amsfonts,bm}

\def\eqref#1{equation~\ref{#1}}

\def\1{\bm{1}}

\def\vh{{\bm{h}}}

\def\vk{{\bm{k}}}

\def\vr{{\bm{r}}}

\def\vv{{\bm{v}}}

\def\vx{{\bm{x}}}

\def\mR{{\bm{R}}}

\def\mW{{\bm{W}}}

\DeclareMathAlphabet{\mathsfit}{\encodingdefault}{\sfdefault}{m}{sl}
\SetMathAlphabet{\mathsfit}{bold}{\encodingdefault}{\sfdefault}{bx}{n}

\newcommand{\R}{\mathbb{R}}

\title{LLM Braces: Straightening Out LLM Predictions with Relevant Sub-Updates}

\author{Ying Shen \\
  Computer Science Department \\
  University of Illinois Urbana-Champaign \\
  \texttt{ying22@illinois.edu} \\\And
  Lifu Huang \\
   Computer Science Department \\
  UC Davis \\
  \texttt{lfuhuang@ucdavis.edu} \\
}

\begin{document}
\maketitle

\begin{abstract}

Recent findings reveal that much of the knowledge in a Transformer-based Large Language Model (LLM) 
is encoded in its feed-forward (FFN) layers, where each FFN layer can be interpreted as the summation of sub-updates, each corresponding to a weighted column vector from the FFN's value parameter matrix that often encodes human-interpretable concepts. 
In light of this, we hypothesize that model performance and behaviors can be further enhanced and controlled by
modulating the contributions of these sub-updates based on their relevance to the input or target output style,  
and propose \methodname{}, a novel and efficient method that computes relevance scores associated with value vectors in FFN layers and leverages these scores to dynamically adjust the contribution of sub-updates. 
By optimizing sub-update contributions, \methodname{} refines the prediction process, leading to more accurate and reliable outputs, much like a `brace' providing support and stability.
Moreover, \methodname{} can be extended to support conditional control over generation characteristics, such as sentiment, thereby offering fine-grained steering of LLM outputs.
Extensive experiments on various LLMs—including Qwen2.5-1.5B, Llama2-7B, and Llama3-8B—demonstrate that \methodname{} outperforms baseline approaches in both fine-tuning and zero-shot settings while requiring significantly fewer tunable parameters, up to 75\% fewer compared to LoRA.
Furthermore, \methodname{} excels in sentiment-controlled generation and toxicity reduction, highlighting its potential for flexible, controlled text generation across applications.

\end{abstract}
\begin{figure}[!t]
    \centering
    \includegraphics[width=\linewidth]{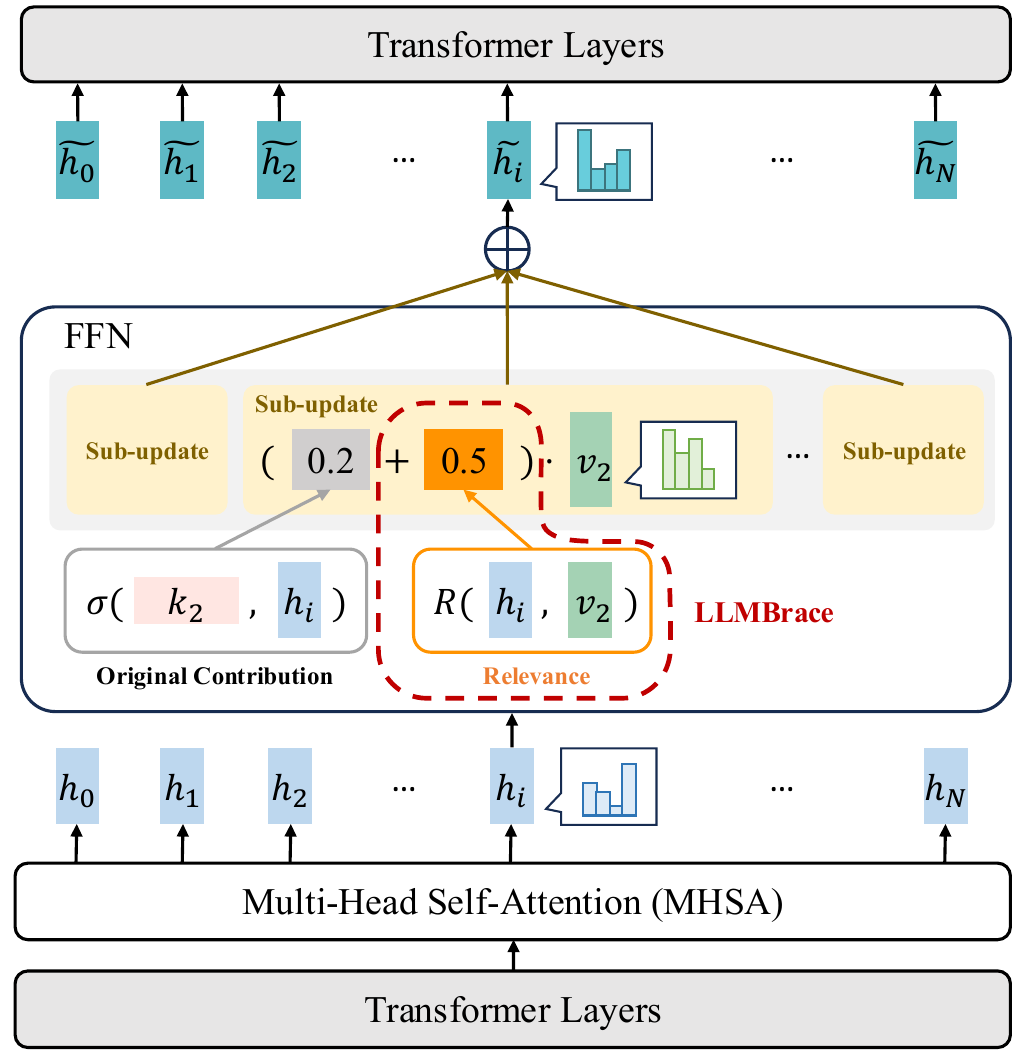}
    \vspace{-6mm}
    \caption{An illustration of feed-forward network (FFN) in a Transformer and its augmentation by \methodname{}. Each FFN's output 
    can be viewed as a sum of \colorbox{sub-updates_bg}{\textbf{sub-updates}}, 
    each corresponding to a weighted FFN \textcolor{value}{value parameter vector}:
    $\textrm{FFN}(\vh_i) = \textcolor{value}{\mW_V} \cdot \sigma \left(\textcolor{key}{\mW_K} \cdot \vh_i\right) = \sum_{j=1}^{d_m} \colorbox{sub-updates_bg}{$\displaystyle \sigma\left(\vh_i \cdot \textcolor{key}{\vk_j}\right) \cdot \textcolor{value}{\vv_j}$}$.
    \methodname{} adjusts the weight of each value parameter vector based on its relevance to the input or target output style, using a relevance function $R(\cdot)$, to modulate model predictions and behaviors.
    }
    \label{fig:teaser}
\end{figure}

\section{Introduction}

Transformer-based Large Language Models (LLMs)~\citep{devlin2019bert,radford2019language,brown2020language,black2022gpt,touvron2023llama,jiang2023mistral,anthropic2023claude} have demonstrated impressive capabilities across a range of natural language processing tasks, yet their internal workings—specifically how they generate predictions—remain somewhat elusive. 
Recent efforts have focused on demystifying the LLMs through interpreting their neurons~\citep{dai2022knowledge,voita2023neurons}, attention mechanisms~\citep{geva2023dissecting,nanda2023factfinding,ferrando2024information}, as well as internal parameters~\citep{geva2022transformer,dar2023analyzing}.

One of the main findings in previous studies is that knowledge is primarily stored in the feed-forward (FFN) layers of Transformer-based language models~\citep{dai2022knowledge,geva2021transformer,geva2022transformer,yao2024knowledge}. \citet{geva2021transformer,geva2022transformer} identify each FFN layer as a key-value memory—
each ``key'' is associated with specific textual patterns from the training data, while each ``value'' represents a distribution over the output vocabulary. Consequently, an FFN update can be decomposed into a collection of \textit{sub-updates}, where each sub-update is represented by a weighted  ``value'' parameter vector within the FFN, as shown in \cref{fig:teaser}. These value vectors often encode human-interpretable concepts, with the associated weights determining the contribution of each sub-update. From the perspective of the residual stream~\citep{elhage2021mathematical}, these sub-updates can be understood as making incremental additive contributions to the model's final output distribution.

Building on these insights, 
we hypothesize that model accuracy can be improved by dynamically modulating 
sub-updates based on the relevance of value vectors to the input—suppressing less relevant ones to reduce noise while amplifying highly aligned ones to enhance predictions. 
Based on this motivation, we introduce \methodname{}, a novel approach that 
employs a learnable relevance module to project both hidden states and value vectors into a common low-dimensional subspace using a low-rank projection with orthonormal rows, which not only ensures computational efficiency but also faithfully preserves the geometric structure of the original high-dimensional representations~\cite{halko2011finding}.
In this transformed space, we compute relevance scores that quantify the alignment between each value vector and the input context, and combine them 
with the original contribution weights through a learnable gating mechanism, which dynamically adjusts the influence of each sub-update, enhancing the model prediction process. 
Moreover, \methodname{} can be easily extended to
steer model outputs toward specific attributes, e.g., a desired sentiment or style, 
by adapting the relevance module to  
measure the alignment between value vectors and attribute-specific token representations.
This conditional relevance score is then integrated 
into the overall relevance score, with an adjustable scalar parameter that allows users to control the influence of this attribute-specific signal during inference.

We validate the effectiveness of \methodname{} through extensive experiments on a variety of LLMs, including Qwen2.5-1.5B~\citep{yang2024qwen2}, Llama2-7B~\citep{touvron2023llama}, and Llama3-8B~\citep{dubey2024llama}.
 Our evaluations encompass both fine-tuning performance on commonsense reasoning tasks and zero-shot performance across a range of tasks designed to assess the models' capabilities in factual knowledge, hallucination reduction, and overall performance on aggregated natural language benchmarks. Notably, \methodname{} improves the average performance by 13.9\% on Qwen2.5-1.5B, 19.6\% on Llama2-7B, and 29.7\% on Llama3-8B compared to LoRA in the zero-shot setting.
 Our results indicate that \methodname{} improves the trustworthiness of LLM and demonstrate its effectiveness in retaining knowledge and facilitating reasoning in the zero-shot setting.
Furthermore, we assess \methodname{} in the context of conditional text generation. In particular, our method proves effective in sentiment steering and toxic language suppression tasks, achieving strong performance in both fine-tuning and zero-shot settings.

In summary, our contributions are threefold: 
1) We propose \methodname{}, a novel and efficient method that computes and leverages relevance scores associated with value vectors to dynamically adjust sub-updates in FFN layers.
2) We extend \methodname{} to support task-specific controlled generation, offering a flexible framework for steering LLM outputs in practical applications.
3) We perform extensive experiments demonstrating that our approach improves model performance—enhancing factual accuracy, improve trustworthiness, and enabling controlled text generation across multiple LLMs and tasks.

\section{Related Work}

\paragraph{Understanding LLMs' Internal Working}

As large language models (LLMs) grow in sophistication, 
research has sought to uncover their internal mechanisms to better understand their 
generation processes. 
Recent studies have explored projecting internal representations onto model's vocabulary~\citep{nostalgebraist_2020,geva2022transformer,katz2024backward,belrose2023eliciting}, applying targeted interventions in transformer computation~\citep{finlayson2021causal,stolfo2023mechanistic,ghandeharioun2024patchscope,geiger2024finding}, and investigating the computation graph of language models~\citep{wanginterpretability,merullocircuit,yao2024knowledge}. %
These studies offer 
valuable insights into how LLMs encode and retrieve knowledge. 
Building on this, our work revisit the role of feed-forward networks in representing knowledge~\citep{geva2021transformer,geva2022transformer} and leverages this insight to develop a simple, parameter-efficient approach for more robust and controllable LLM generation.

\vspace{-1mm}
\paragraph{Model Editing} 
Model editing~\citep{sinitsin2019editable} 
enables fast, targeted updates to a pre-trained model's behavior without full retraining, 
allowing it to adapt to new information or correcting specific errors efficiently. 
Recent work on LLMs has focused on modifying stored knowledge to update or correct factual outputs~\citep{meng2022locating,mengmass,huangtransformer}. Methods include integrating auxiliary networks~\citep{huangtransformer,hartvigsen2024aging} or directly modifying model parameters responsible for specific outputs~\citep{dai2022knowledge,meng2022locating,mengmass}. 
Unlike these studies which focused on changing particular outputs for certain model inputs, 
our work seeks to enable more flexible and generic updates to model behavior, rather than focusing on particular aspects.

\vspace{-1mm}
\paragraph{Controllable Text Generation}
Controllable text generation~\cite{zhang2023survey} aims to guide 
Existing methods include fine-tuning models for specific constraints~\cite{keskar2019ctrl,he2021parallel}, prompt engineering~\cite{reif-etal-2022-recipe,zhou2023controlled}, and latent space manipulation~\cite{zou2023representation,liucontext,turner2023activation,han2024word}. 
Unlike these approaches that focus solely on conditional generation, our method dynamically modulates sub-update contributions based on their relevance to the input or target output style. This not only enables more effective conditional generation but also enhances overall model performance on general tasks by refining the prediction process.

\section{Preliminaries}
\label{sec:background}

\begin{figure*}[!t]
    \centering
    \includegraphics[width=\linewidth]{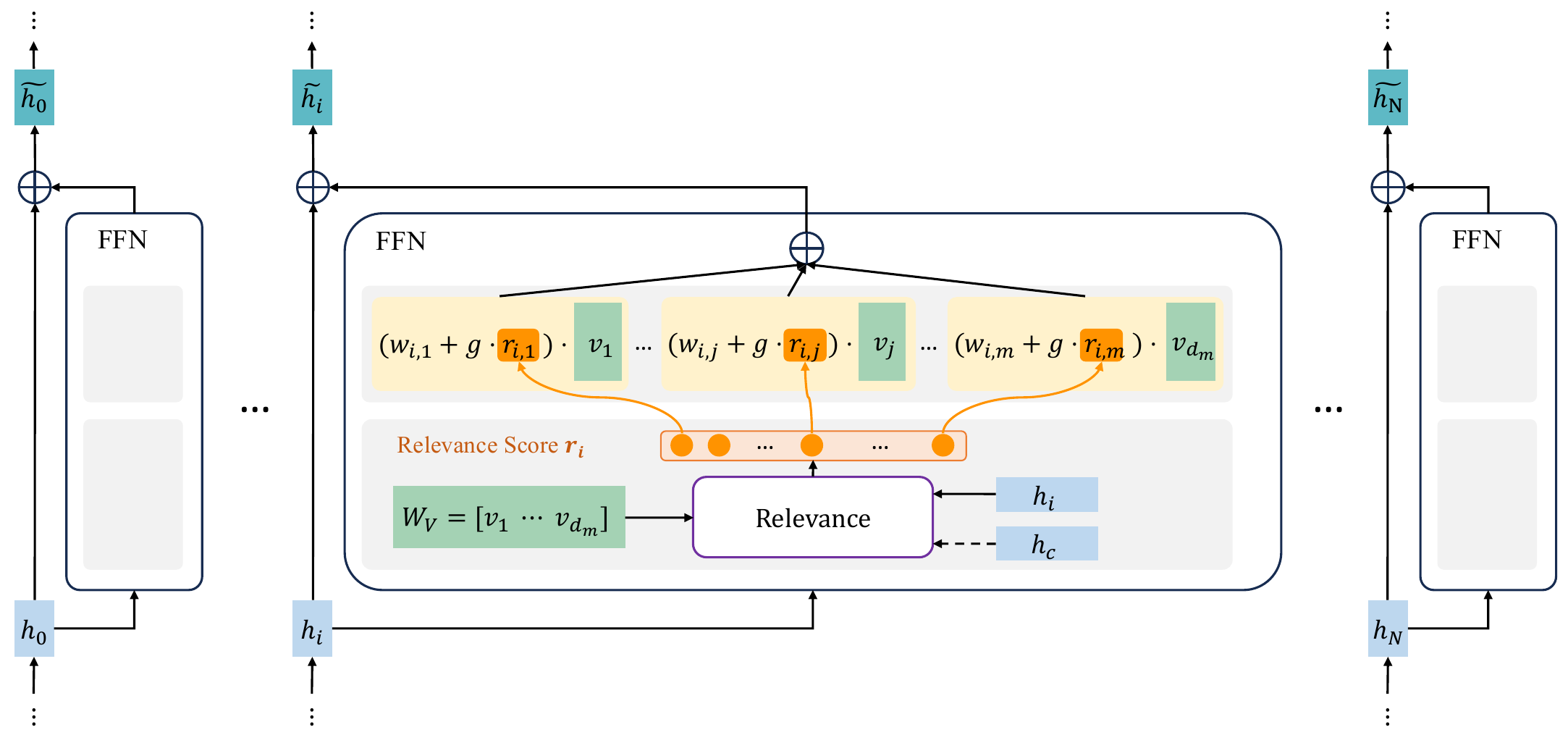}
    \vspace{-2mm}
    \caption{Overview of \methodname{}. 
     The relevance module computes scores that assess the alignment between each value parameter vector $\vv_j$ and the input context $\vh_i$. For conditional generation, an additional branch computes the conditional relevance scores with respect to task-specific attribute representations $\vh_c$.
     These relevance scores are then combined with the original contribution weights to dynamically modulate the FFN sub-updates.
    }
    \label{fig:overview}
\end{figure*}

In this work, we mainly focus on Transformer-based large language models which have shown exceptional performance across various tasks. Given a sequence of input tokens $x = [x_0, \dots, x_N]$, 
a transformer-based LMM first encodes the tokens into vectors representations $\vh^0 = [\vh^0_0, \cdots, \vh^0_N] \in \mathbb{R}^{N \times d}$,\footnote{Bold symbols denote vectors throughout this paper.} 
which are then iteratively updated through a sequence of $L$ Transformer layers, 
each consisting of a multi-head self-attention (MHSA) sublayer followed by a feed-forward (FFN) sublayer, interconnected by residual connections.

\paragraph{Feed-Forward Layer (FFN)}
As shown in \cref{fig:teaser}, each FFN layer $l$ takes the hidden state output $\vh^{\ell}_i$ at the $i$-th token position from its preceding $l$-th MHSA layer as input, then applies a transformation, resulting in an updated representation $\tilde{\vh}^{\ell}_i$ via residual connection:
\begin{equation}
\small
    \tilde{\vh}^{\ell}_i = \textrm{FFN}^{\ell}(\vh^{\ell}_i) + \vh^{\ell}_i = \mW^{\ell}_V \cdot \sigma\left(\mW^{\ell}_K \cdot \vh^{\ell}_i \right) + \vh^{\ell}_i, 
\end{equation}
where $\mW^{\ell}_V \in \mathbb{R}^{d \times d_m}$ and $\mW^{\ell}_K \in \mathbb{R}^{d_m \times d}$ can be conceptualized as the key and value parameter matrices with inner-dimension $d_m$ at layer $\ell$, and $\sigma(\cdot)$ denotes a nonlinear activation function. 

Based on~\citet{geva2021transformer,geva2022transformer}, the $\ell$-th FFN layer can be further dissected into:
\begin{align}
    \textrm{FFN}^{\ell}(\vh^{\ell}_i) = \sum_{j=1}^{d_m} w^{\ell}_{i,j} \cdot \vv^{\ell}_j,
    \label{eq:sum_update}
\end{align}
where each term $(w^{\ell}_{i,j} \cdot \vv^{\ell}_j)$ acts as a sub-update within the layer and $w^{\ell}_{i,j} = \sigma(\vh^{\ell}_i \cdot \vk^{\ell}_j)$ is the coefficient that assigns weights to the corresponding value vector $\vv^{\ell}_j$. Here, $\vk^{\ell}_j$ and $\vv^{\ell}_j$ denote the $j$-th row of $\mW^{\ell}_K$ and the $j$-th column of $\mW^{\ell}_V$, respectively. This decomposition reveals that each FFN update can be viewed as an aggregation of these sub-updates, where each sub-update is effectively a weighted value vector. These sub-updates incrementally contribute to the model's final output distribution.

\section{\methodname{}}

Building on insights into FFNs, we introduce \methodname{} to dynamically modulate the contributions of sub-updates in each FFN layer, 
aiming to efficiently refine the model prediction process and make it more accurate and controllable. Notably, \methodname{} introduces only a minimal number of additional parameters and can be seamlessly integrated into any existing transformer-based LLMs. \cref{fig:overview} shows an overview of \methodname{}.

\subsection{Relevance of Sub-updates}

As shown in \cref{eq:sum_update}, each sub-update $(w^{\ell}_{i,j} \cdot \vv^{\ell}_j)$ in FFN represents a weighted \textit{value} parameter vector, which usually corresponds to a set of human interpretable concepts.
Given this, we postulate that the accuracy of the model's predictions can be improved by increasing the contribution of value vectors that are highly relevant to the given input while minimizing the contribution of those that are not. 
To achieve this, we propose to explicitly measure the relevance of each value vector to the current input, i.e., the hidden states of input tokens, which provide rich contextual information for the input sequence. 
To this end, we introduce a \textbf{Relevance Module} $R(\cdot)$ that computes a relevance score $\vr^{\ell}_i \in \R^{d_m}$ for each FFN layer $\ell$ and token position $i$:
\begin{equation}
\small
    \vr^{\ell}_i = R(\mW^{\ell}_V, \vh^{\ell}_{i}) = ( \mR^{\ell} \mW_V^{\ell})^\top (\mR^{\ell} \vh^{\ell}_{i}) / \sqrt{d_r},
    \label{eq:rel}
\end{equation}
where $\mW^{\ell}_V \in \R^{d \times d_m}$ denotes the value parameter matrices at layer $\ell$, $\vh_{i} \in \R^{d}$ is the hidden states for token $i$, and $\mR^{\ell} \in \R^{d_r \times d}$ is a learnable low-rank projection matrix with orthonormal rows. Here, $d$ is the dimensionality of hidden states and $d_r < d$ is the dimensionality of rank of the subspace.
The use of a low-rank projection with orthonormal rows ensures that the relevance scores are computed efficiently while faithfully preserving the geometric structure of the original high-dimensional representations~\cite{halko2011finding}, providing a reliable foundation for subsequent augmentation.

The resulting vector $\vr^{\ell}_i = [\vr^{\ell}_{i,0}, \dots, \vr^{\ell}_{i,d_m}]^\top$ comprises individual relevance scores, with each component $\vr^{\ell}_{i,j}$ corresponds to the $j$-th sub-update $(w^{\ell}_{i,j} \cdot \vv^{\ell}_j)$. 
In effect, the relevance score measures how well each sub-update aligns with the input context, allowing us to modulate the contribution of sub-updates based on their relevance to improve model performance.

\subsection{Relevance-based Augmentation}

Armed with the relevance scores, we refine the FFN layers by incorporating them as an additive adjustment to the contribution of each sub-update. 
The standard FFN layer at $\ell$-th layer (\cref{eq:sum_update}) is augmented as follows:
\begin{align}
    \textrm{FFN}_{\textrm{AUG}}^{\ell}(\vh^{\ell}_i) = \sum_{j=1}^{d_m} \left(w^{\ell}_{i,j} + g^{\ell} \cdot r^{\ell}_{i,j}\right) \cdot \vv^{\ell}_j,
    \label{eq:aug}
\end{align}
where $r^{\ell}_{i, j}$ is the relevance score associated with the value vector $\vv^{\ell}_j$ relative to the hidden states $\vh^{\ell}_i$
and $g^{\ell}$ is the learnable gating mechanism at layer $\ell$ that adaptively controls the strength of relevance augmentation. 
We set $g^{\ell} = \sigma(g)$ where $\sigma(\cdot)$ is sigmoid activation function, and initialize $g$ to a small value (e.g. $g = -5$) to ensure that the gating mechanism starts nearly at zero. The near zero-initialization supports stable early-stage training and allows the model to progressively incorporate relevance-based augmentation as training proceeds. 

We choose the additive formulation because of its simplicity and its ability to handle cases where the original contribution score $w^{\ell}_{i,j}$ is negative — something that binary filtering or relevance-based multiplication may not accommodate as effectively. 
By adding the gated relevance score, \methodname{} dynamically amplifies sub-updates that are aligned with the input context while suppressing those that may introduce noise or errors. During training, we adopt the standard language modeling objectives. We freeze the weights of the original language backbone model and only fine-tune the newly introduced parameters.

\subsection{Task-Specific Relevance}

Thus far, we have focused on leveraging the model's hidden states to compute relevance scores, refining FFN sub-updates to improve prediction accuracy and reliability. 
\methodname{} can also be extended for controlled text generation, steering outputs toward desired attributes such as sentiment.

To enable attribute-based control, we introduce a conditional relevance score, $\vr_c$, which quantifies the relevance between the value parameter vectors and attribute-specific tokens. 
For instance, if we wish to generate text with a positive sentiment, the attribute-specific tokens might include words such as ``happy'', ``joyful'', etc. 
To compute $\vr_c$, we first pass the attribute-specific tokens through the target language model and extract their hidden states over $N$ tokens at each layer $\ell$, resulting in $\tilde{\vh}^{\ell}_c \in \R^{N \times d}$.
We then apply average pooling on $\tilde{\vh}^{\ell}_c$ to obtain one aggregated representation, followed by multi-layer perceptron (MLP) projection $f_c(\cdot)$:
\begin{align}
    \vh^{\ell}_c = f_c\Big(\text{Pooling}(\tilde{\vh}^{\ell}_c)\Big) \in \R^{d}.
\label{eq:pool}
\end{align}

Once we have the projected attribute-specific representation $\vh^{\ell}_c$ at each layer $\ell$, we compute the conditional relevance score $\vr^{\ell}_{c}$ in a manner analogous to the primary relevance score:
\begin{equation}
\small
    \vr^{\ell}_{c} = R(\mW^{\ell}_V, \vh^{\ell}_{c}) = ( \mR^{\ell} \mW_V^{\ell})^\top (\mR^{\ell} \vh^{\ell}_{c}) / \sqrt{d_r}
\label{eq:rel_c}
\end{equation}
where $\mR^{\ell} \in \R^{d_r \times d}$ is the same learnable low-rank projection matrix with orthonormal rows in \cref{eq:rel}.

To integrate this conditional relevance into the overall model, we update the original relevance score (\cref{eq:rel}) as follows:
\begin{align}
    \vr^{\ell}_{i} \leftarrow \vr^{\ell}_{i} + s \cdot \vr^{\ell}_{c},
\label{eq:rel_update}
\end{align}
where $s$ denotes an adjustable scalar indicating the polarity and the intensity of the conditional relevance. 
We choose a scalar rather than a learnable gating function so that users can manually control the degree to which the model’s outputs are influenced by the desired attribute during inference. 

The incorporation of task-specific relevance enables \methodname{} not only to improve prediction accuracy by emphasizing relevant sub-updates but also to guide the generation process in accordance with specific task attributes.
\section{Experimental Setup}

\paragraph{Backbone Models and Datasets}
We implement \methodname{} on three popular large language models: Qwen2.5-1.5B~\citep{yang2024qwen2}, Llama2-7B~\citep{touvron2023llama}, and Llama3-8B~\citep{dubey2024llama}, and evaluate the models under three experimental settings: (1) \textbf{Supervised Fine-tuning}, where we follow \cite{hu2023llm} and evaluate all methods on Commonsense Reasoning datasets. 
Specifically, we leverage Commonsense170K~\cite{hu2023llm}, which is a combination of eight commonsense reasoning datasets, including BoolQ~\cite{clark2019boolq}, PIQA~\cite{bisk2020piqa}, SIQA~\cite{sap2019social}, HellaSwag~\cite{zellers2019hellaswag},
WinoGrande~\cite{sakaguchi2021winogrande}, ARC-e, ARC-c~\cite{clark2018think}, and OBQA~\cite{mihaylov2018can}, for both training and evaluation. (2) \textbf{Zero-Shot Generalization}, where we first tune \methodname{} on Dolly~\cite{DatabricksBlog2023DollyV2}, an instruction-following dataset containing 15K examples, and evaluate \methodname{} across multiple datasets targeting different aspects of model behavior. 
For factual knowledge, we use PopQA~\citep{mallen-etal-2023-trust}, TriviaQA~\citep{joshi2017triviaqa}, and Natural Questions (NQ)~\citep{kwiatkowski-etal-2019-natural}. To assess truthfulness and trustworthiness of LLMs, we employ TruthfulQA~\citep{lin2022truthfulqa}.
Additionally, we also test all methods on AGI Eval~\citep{zhong2024agieval}, an aggregated benchmark dataset designed to assess overall model capabilities. (3) \textbf{Conditional Generation}, where we follow the setting in ~\citet{han2024word}, using Stanford Sentiment Treebank (SST5)~\citep{socher2013recursive} for sentiment steering and Jigsaw Unintended Bias in Toxicity Classification~\citep{jigsaw-unintended-bias-in-toxicity-classification} for toxic suppression.

\begin{table*}[!t]
  \centering
  \resizebox{\linewidth}{!}{%
  \begin{tabular}{c | l r r | c c c  c c c  c c c c}
  \toprule
    & \textbf{Method} & Param. & Param. (\%) & BoolQ & PIQA & SIQA & HellaS. & WinoG. & ARC-e &ARC-c & OBQA & AVG \\ 

   \midrule
  \multirow{4}{0.6cm}{\rotatebox{90}{\shortstack{Qwen2.5-\\1.5B}}} & LoRA (r=16) & 2.2M & 0.14 & 66.91	&80.85	&75.64&	\textbf{89.99}&	\textbf{76.47}	&89.48&	74.48&	\textbf{84.80}&	79.83\\
  & LoRA (r=32) & 4.4M & 0.28 & 66.67	&82.80&	75.69&	89.52	&76.40&	88.01	&77.04&	83.20&	79.92 \\ \cline{2-13}
  & \methodname{} (r=16) & 0.6M  & 0.04 & 66.36&	82.59	&\textbf{75.84}&	89.41&	76.32&	89.10&	\textbf{78.58}&	84.00&	80.28 \\
  & \methodname{} (r=32) & 1.2M  & 0.08 & \textbf{67.49}&	\textbf{83.03}	&75.38	&89.90&	76.32&	\textbf{89.69}	&77.13	&83.60&	\textbf{80.32} \\
   \midrule
   \multirow{4}{0.6cm}{\rotatebox{90}{Llama2-7B}} &LoRA (r=16) & 8.4M & 0.12 &\textbf{71.62}&\textbf{84.71} &78.86 & 93.19 &82.40 &\textbf{85.98} &71.08 &\textbf{84.80}&\textbf{81.58}\\
  & LoRA (r=32) &  16.8M & 0.25 & 70.52&	82.69&	79.11&	92.00&	83.74	&84.42	&68.94	&82.80&	80.53\\ \cline{2-13}
  & \methodname{}  (r=16)  & 2.1M & 0.03 & 71.04&	83.41	&\textbf{80.14}&	\textbf{93.40}&	\textbf{84.21}&	85.48&	\textbf{71.33}&	83.00&	81.50  \\
  & \methodname{} (r=32) & 4.2M & 0.06  & 70.61	&83.95	&80.09	&93.27	&82.79&	85.90&	70.99&	82.00&	81.20  \\
  \midrule
    \multirow{4}{0.6cm}{\rotatebox{90}{Llama3-8B}} & LoRA (r=16) & 6.8M & 0.08 & 73.27&	88.25&	78.92&	94.96&	\textbf{87.21}&	90.95&	77.99	&86.20&	84.72\\
    & LoRA (r=32) & 13.6M  & 0.17 & 71.31	&85.36&	79.83& 92.97	&83.43&	87.12&	74.15&	84.00&	82.27\\ \cline{2-13}
    & \methodname{}  (r=16)  & 2.1M & 0.03 & 72.75	&88.52	&81.17&	95.27&	86.42	&92.59	&80.97&	87.80&	85.69  \\
  & \methodname{} (r=32) & 4.2M & 0.05 & \textbf{74.40}&	\textbf{89.12}	&\textbf{81.37}&	\textbf{95.55}&	86.66	&\textbf{92.97}&	\textbf{83.19}&	\textbf{88.80}&	\textbf{86.51}\\
  \bottomrule
  \end{tabular}
  }
  \caption{Accuracy comparison on eight \textit{commonsense reasoning} tasks across different LLM architectures.  Param. indicates the number of tunable parameters and Param. (\%) is calculated by dividing the number of trainable parameters by the number of parameters of the base LM.}
  \label{table:commonsense}
  \end{table*}
\begin{table*}[!t]
  \centering
  \resizebox{\linewidth}{!}{%
  \begin{tabular}{c | l r r | c c c   c  c c c }
  \toprule
    & \textbf{Method} & Param. & Param. (\%) &  TruthfulQA (MC1) &  TruthfulQA (MC2) & PopQA  & TriviaQA & NQ & AGI. & AVG \\ 

   \midrule
  \multirow{4}{0.6cm}{\rotatebox{90}{\shortstack{Qwen2.5-\\1.5B}}} & LoRA (r=16) & 2.2M & 0.14 &  24.60 &	38.41&	17.87&		26.47	&4.76&	30.43 & 23.76\\
  & LoRA (r=32) & 4.4M & 0.28 & \textbf{26.19}&	39.24&	17.66	&	24.95	&5.29&	29.18 & 23.75 \\ \cline{2-11}
  & \methodname{} (r=16) & 0.6M  & 0.04 & 26.07&	39.74&	17.80&	35.40&	9.50&	31.67 & 26.70\\
  & \methodname{} (r=32) & 1.2M  & 0.08 & 26.07	&\textbf{39.93}	&\textbf{18.44}	&\textbf{36.26}&	\textbf{9.58}	&\textbf{32.11} & \textbf{27.07}\\
   \midrule
    \multirow{4}{0.6cm}{\rotatebox{90}{Llama2-7B}}& LoRA (r=16) & 8.4M & 0.12 & 25.21&	39.24	&25.27	&	43.93	&8.25	&21.13 & 27.17\\
  & LoRA (r=32) &  16.8M & 0.25 & 24.36&	38.75&	26.08	&40.34&	8.67&	22.66 & 26.81 \\ \cline{2-11}
  & \methodname{}  (r=16)  & 2.1M & 0.03 & \textbf{26.93}&	\textbf{40.56}	&28.14&	\textbf{55.20}&	\textbf{20.30}&	\textbf{23.83}&	\textbf{32.49} \\
  & \methodname{} (r=32) & 4.2M & 0.06  & 25.95&	38.89&	27.20	&52.98&	16.62&	23.94	&30.93\\
  \midrule
    \multirow{4}{0.6cm}{\rotatebox{90}{Llama3-8B}} & LoRA (r=16) & 6.8M & 0.08 & 27.42&	42.62	&32.40&	34.04&	11.14	&24.61 & 28.71 \\
    & LoRA (r=32) & 13.6M  & 0.17 & \textbf{29.62} &	43.14&	32.06&	23.15&	9.11&	25.80 & 27.15 \\ \cline{2-11}
    & \methodname{}  (r=16)  & 2.1M & 0.03  & 28.27	&\textbf{43.91}&	\textbf{36.21}	&\textbf{66.11}&	\textbf{19.53}&	\textbf{29.47} & \textbf{37.25} \\
  & \methodname{} (r=32) & 4.2M & 0.05 & 28.52	&43.46&	36.08&	64.56&	19.25&	29.23 & 36.85 \\
  \bottomrule
  \end{tabular}
  }
  \caption{Zero-shot comparison on various tasks across different LLM architectures.  Param. indicates the number of tunable parameters and Param. (\%) is calculated by dividing the number of trainable parameters by the number of parameters of the base LM.}
  \label{table:zero_shot}
  \end{table*}

\paragraph{Baselines}
For both supervised fine-tuning and zero-shot generalization, we compare \methodname{} with LoRA~\cite{hulora}, a parameter-efficient fine-tuning method that uses low-rank matrices to approximate additive weight updates during training.
For conditional generation, we consider the following baselines: DExperts~\citep{liu2021dexperts}, GeDi~\citep{krause2021gedi}, PromptT5~\citep{raffel2020exploring}, LoRA~\cite{hulora} and LMSteer~\citep{han2024word}. 
These baselines employ various techniques, including classifier-guided decoding, Bayesian rule-based generation, prompt-based conditioning, parameter-efficient adaptation, and embedding steering, to steer language models' outputs. More details on these baselines can be found in \cref{app:baselines}.

\paragraph{Implementation Details}
For supervised fine-tuning, we conduct hyperparameter tuning for both the baseline and different variants of \methodname{}. Specifically, we randomly sample 300 instances from the training split of Commonsense170K to optimize hyperparameters for commonsense reasoning. The optimal hyperparameters are selected based on performance on held-out validation sets.
For zero-shot generalization, we use the same set of hyperparameters selected for Commonsense Reasoning.
Further details on the hyperparameter settings for each task are provided in the \cref{app:hyper}.

\section{Results and Discussion}

\subsection{Supervised Fine-tuning Performance}

As shown in \cref{table:commonsense}, for supervised fine-tuning performance on commonsense reasoning tasks, 
\methodname{} achieves competitive, and even superior, performance while requiring significantly fewer trainable parameters.
For instance, on the Llama3-8B model, \methodname{} outperforms LoRA by 5.2\% while reducing the number of trainable parameters by approximately 38\%.
These results demonstrate that \methodname{} can efficiently enhance the model performance and reliability, while matching or surpassing the strong model fine-tuning based baselines, with greater parameter efficiency.

\begin{table*}[!t]
  \centering
  \resizebox{\linewidth}{!}{%
  \begin{tabular}{l | l | ccccc c ccccc}
  \toprule
   & \multirow{2}{*}{\centering \textbf{Method}}   & \multicolumn{5}{c}{\textbf{Neutral to Positive}}  && \multicolumn{5}{c}{\textbf{Neutral to Negative}}\\
   \cline{3-7} \cline{9-13}
  &&   Positivity (\%) $\uparrow$ & PPL $\downarrow$ & Dist-1 $\uparrow$ & Dist-2 $\uparrow$ & Dist-3 $\uparrow$ && Positivity $\downarrow$ & PPL $\downarrow$ & Dist-1 $\uparrow$ & Dist-2 $\uparrow$ & Dist-3 $\uparrow$ \\ \cline{2-13}
  & GPT2-Large & 53.55 & 12.48 & 0.52 & 0.82 & \textbf{0.85} && 53.55 & 12.48 & 0.52 & 0.82 & \textbf{0.85} \\
  \midrule
  \multirow{5}{0.6cm}{\rotatebox{90}{\shortstack{Supervised FT}}} & LoRA~\cite{hulora} & 26.88 & 158.56 & \textbf{0.57} &  0.82 & 0.83 &&  20.08 &192.13& 0.55 &0.78 &0.79 \\
  &DExperts~\citep{liu2021dexperts} &\textbf{94.46} & 45.83 &0.56& \textbf{0.83}& 0.83 && \textbf{3.77}& \textbf{45.91}& 0.60& \textbf{0.84} &0.83 \\
  &GeDi~\citep{krause2021gedi} & 86.01 & 58.41& 0.57 &0.80& 0.79 &&  8.73& 84.11& \textbf{0.63}& 0.84& 0.82\\
  &LMSteer~\citep{han2024word} & 90.70& \textbf{41.20}& 0.46& 0.78 &0.83 &&  8.02& 57.74 &0.48& 0.78 &0.80 \\ \cline{2-13}
  & \methodname{} (r=16)   & 91.20 & 52.63 & 0.38 & 0.73 & 0.81 &&  7.10 & 63.49 & 0.42 & 0.75 & 0.83  \\
  \midrule
  \multirow{2}{0.6cm}{\rotatebox{90}{\shortstack{Zero\\-Shot}}} &PromptT5~\citep{raffel2020exploring} & 68.12 & 37.30& \textbf{0.58}& \textbf{0.78} &0.72 && 25.78 &48.60 &\textbf{0.60}& 0.78 &0.70 \\ \cline{2-13}
  & \methodname{} (r=16)   & \textbf{69.82} & \textbf{30.03} & 0.46 & 0.77 & \textbf{0.85} && \textbf{12.96} & \textbf{39.78} & 0.46 & \textbf{0.79} & \textbf{0.85} \\ %
  \bottomrule
  \end{tabular}}
  \caption{Sentiment steering performance. Results of all baseline
methods are taken from ~\citet{han2024word}.}
  \label{table:sent}
  \end{table*}  

\begin{table}[!t]
  \centering
  \resizebox{\linewidth}{!}{%
  \begin{tabular}{l | c | p{0.7\linewidth} }
  \toprule
   Sentiment & Steering Value  & Generation \\
   \midrule
   \multirow{9}{*}{Positive} & \multirow{3}{*}{1}  & \underline{It is for this reason that I believe our} film has more of a chance of being good than bad.\\ \cline{2-3}
    & \multirow{3}{*}{3} & \underline{It is for this reason that I believe our}  film will leave a lasting impression upon you.\\ \cline{2-3}
    & \multirow{3}{*}{5} & \underline{It is for this reason that I believe our} film is one of the year 's most inspiring and accessible films. \\
   \midrule
   \multirow{9}{*}{Negative} & \multirow{3}{*}{1} & \underline{It is for this reason that I believe our} film 's title should have been something else. \\ \cline{2-3}
    & \multirow{3}{*}{3} & \underline{It is for this reason that I believe our} film is going to be overlooked by the viewing public. \\ \cline{2-3}
    & \multirow{3}{*}{5} & \underline{It is for this reason that I believe our} film is not only mediocre , but also downright ugly and unlikable . \\
  \bottomrule
  \end{tabular}
  }
  \caption{Qualitative examples of sentiment steering using \methodname{}. \underline{Underlined text} indicates the input provided to the model, while the remaining text is generated by \methodname{}.
  }
  \label{table:sent_ex}
  \end{table}

\subsection{Zero-shot Generalization}
\cref{table:zero_shot} presents a comparison of zero-shot performance across various LLM architectures. 
\methodname{} consistently surpasses 
baselines across all tasks while requiring significantly fewer trainable parameters. 
Specifically, it improves the average performance by 13.9\% on Qwen2.5-1.5B, 19.6\% on Llama2-7B, and 29.7\% on Llama3-8B compared to LoRA, highlighting its effectiveness and generalization in enhancing model performance by modulating the subupdates in FFNs. 
\methodname{} excels particularly in knowledge-intensive tasks such as TriviaQA, PopQA, and NQ. These tasks require the model to accurately answer questions that demand factual knowledge, showcasing the effectiveness of \methodname{} in retaining knowledge and facilitating reasoning in the zero-shot setting. 
Moreover, \methodname{} also shows consistent gains on TruthfulQA MC2, indicating its effectiveness in improving truthfulness and reducing hallucination.
All these results confirm our initial hypothesis that modulating the contributions of sup-updates in FFNs enhances the model's prediction process, making it more accurate and reliable.

\subsection{Conditional Generation}

We further demonstrate the versatility of \methodname{} in controlling the attributes of generated text, with experiments focusing on sentiment steering and toxic language suppression. 
We evaluate \methodname{} under both supervised fine-tuning and zero-shot settings.

\paragraph{Supervised Sentiment Steering}
To evaluate \methodname{}'s ability to control sentiment in generated text, 
we first define a set of words representing the target sentiment (e.g., positive or negative), as listed in \cref{tab:attributes}, and concatenate the representative words into a string of attribute-specific tokens for each sentiment, 
which are then used to compute the relevance scores following \cref{eq:pool,eq:rel_c,eq:rel_update}. 
During training, we pair positive tokens with a positive steering value $s=1$ (\cref{eq:rel_update}) and negative tokens with a negative steering value $s=-1$, 
with maximal likelihood as the training objective. 
When negative texts are available, we fit the model by pairing negative texts with positive tokens alongside a negative steering value $s=-1$, or positive tokens with a positive steering value $s=1$. This strategy essentially enlarges the training dataset and enhances the model's robustness. 
During inference, \methodname{} adjusts the steering value $s$ to control the sentiment of generated output, 
offering a flexible and efficient approach for sentiment steering.
For benchmarking, we follow \citet{han2024word} which utilizes 5,000 neutral prompts from OpenWebText~\citep{Gokaslan2019OpenWeb} and generates 25 sentences per prompt with GPT2-large. 
We evaluate the generated text based on its level of positivity, fluency (measured via perplexity), and diversity (Dist-\{1,2,3\}). More details for the evaluation metrics are provided in \cref{app:evaluation}.

\begin{table*}[!t]
  \centering
  \resizebox{\linewidth}{!}{%
  \begin{tabular}{l | l  | cc c c c ccc}
  \toprule
   & \multirow{2}{*}{\centering \textbf{Method}} & \multicolumn{2}{c}{\textbf{Toxicity} $\downarrow$}  && \textbf{Fluency} $\downarrow$ && \multicolumn{3}{c}{\textbf{Diversity} $\uparrow$} \\
   \cline{3-4} \cline{6-6} \cline{8-10}
  &&   Max. toxicity & Toxicity prob. && Output ppl. && Dist-1  & Dist-2  & Dist-3  \\
  \midrule
  \multirow{5}{0.6cm}{\rotatebox{90}{\shortstack{Supervised FT}}} & LoRA~\cite{hulora} & 0.365 & 0.210 &&21.11 &&0.53& 0.85 &0.86 \\
 & DExperts~\citep{liu2021dexperts} & 0.314& 0.128 &&32.41&& 0.58 &0.84& 0.84 \\
  &GeDi~\citep{krause2021gedi}  & 0.363& 0.217&& 60.03 &&0.62& 0.84&0.83\\
  &LMSteer~\citep{han2024word}  &  0.249 & 0.089 && 28.26  && 0.55 &  0.84 & 0.84 \\ \cline{2-10}
  &\methodname{} (r=8) & \textbf{0.215} & \textbf{0.061} && \textbf{19.44} && 0.54 & \textbf{0.85} & \textbf{0.86}   \\ \midrule
  \multirow{2}{0.6cm}{\rotatebox{90}{\shortstack{Zero\\-Shot}}} & PromptT5~\citep{raffel2020exploring} & 0.320& 0.172 &&55.10&& \textbf{0.58}& 0.76 &0.70\\ \cline{2-10}
  &\methodname{} (r=16) & \textbf{0.172} & \textbf{0.027} && \textbf{31.91} && 0.51 & \textbf{0.83} & \textbf{0.86}  \\
  \bottomrule
  \end{tabular}}
  \caption{ Toxic Language Suppression. Results of all baseline methods are taken from ~\citet{han2024word}.} 
  \label{table:toxic}
  \end{table*}

\cref{table:sent} shows 
that \methodname{} successfully steers the generated text with the desired sentiment, while achieving a reasonable balance of fluency and diversity.
Additionally, \methodname{} enables control over the intensity of sentiment expression. By controlling the steering value $s$, we can modulate the strength of sentiment in the generated output. 
\cref{table:sent_ex} illustrates how varying the steering value $s$ adjust the intensity of sentiment in the generated text, highlighting \methodname{}'s capability to modulate the emotional tone of the output in a fine-grained manner.

\paragraph{Supervised Toxic Language Suppression}
To mitigate toxic language in generated text, we construct attribute-specific tokens $\vx_c$ using the toxic words from WORDFILTER~\citep{gehman-etal-2020-realtoxicityprompts}. 
These tokens are used to compute the conditional relevance score, 
guiding the model away from toxic completions by assigning a negative steering value $s$. During training, we pair toxic texts with a positive steering value ($s=1$) and non-toxic texts with a negative steering value ($s=-1$).
 
Following the evaluation protocol of \citet{han2024word}, we use 10K nontoxic prompts from REALTOXICITYPROMPTS~\citep{gehman-etal-2020-realtoxicityprompts} and randomly generate 25 sentences using GPT2-large. 
During inference,  we apply a steering value $s=-5$ to suppress the toxic language.
Similar to the sentiment task, we assess the generated text based on fluency and diversity, and measure toxicity using the Perspective API~\footnote{https://perspectiveapi.com/}. More details can be found in \cref{app:evaluation}.
As shown in \cref{table:toxic}, \methodname{} significantly reduces the model's propensity for toxic language by 15.8\% while maintaining superior fluency and diversity, 
demonstrating its effectiveness in suppressing toxic language by decreasing the contribution of toxic-associated sub-updates during generation. Qualitative examples can be found in \cref{app:qualitative}.

\paragraph{Zero-shot Conditional Generation}

We further evaluate \methodname{}'s performance in zero-shot conditional generation.
For this, we construct a new dataset, \textit{StyleDataset}, by selecting 10K sentences from OpenWebText~\citep{Gokaslan2019OpenWeb}. 
For each sentence, we 
prompt Llama3.1-8B-instruct~\citep{dubey2024llama} to generate a short list of descriptive attribute words that capture the sentence's style, resulting in a dataset of 10K sentence-attribute pairs. 
The prompting instructions are detailed in \cref{app:style}.
We then train \methodname{} on \textit{StyleDataset} by pairing the attribute tokens with a positive steering value ($s=1$) and maximizing the likelihood of the corresponding sentence. Since \methodname{} is not fine-tuned specifically for sentiment or toxicity tasks, we can evaluate its zero-shot generalization on both tasks. As shown in \cref{table:sent,table:toxic}, \methodname{} consistently outperforms the zero-shot baseline in both sentiment control and toxicity suppression. 
Notably, for toxic suppression, the zero-shot performance of \methodname{} even exceeds that of its fine-tuned counterpart.
These results highlight \methodname{}'s robustness and adaptability in zero-shot conditional generation tasks, eliminating the need for task-specific fine-tuning.

\begin{table}[!t]
  \centering
  \resizebox{\linewidth}{!}{%
  \begin{tabular}{c | l c c }
  \toprule
    & \textbf{Method}  & Commonsense AVG & Zero-shot AVG    \\ 
   \midrule
  \multirow{4}{0.6cm}{\rotatebox{90}{\shortstack{Qwen2.5-\\1.5B}}}  & \methodname{} (r=16)  &	\textbf{80.28} & \textbf{26.70} \\
  & - \; w/o REL & 78.24 & 26.50\\ \cline{2-4}
  & \methodname{} (r=32)  & 	\textbf{80.32} & \textbf{27.07} \\ 
  & - \; w/o REL & 78.20 & 26.62 \\
   \midrule
    \multirow{4}{0.6cm}{\rotatebox{90}{Llama2-7B}} & \methodname{}  (r=16) & \textbf{81.50} & \textbf{32.49}\\
    & - \; w/o REL & 77.46 & 30.59 \\ \cline{2-4}
    & \methodname{} (r=32) & \textbf{81.20} & \textbf{30.93} \\
    & - \; w/o REL & 77.34 & 30.70 \\
  \bottomrule
  \end{tabular}
  }
  \caption{Effect of relevance module comparison.}
  \label{table:update}
  \end{table}

\subsection{Effect of Relevance Module}
To better understand the contribution of our proposed relevance module, we conduct an ablation study by training an additional variant in which, instead of using the computed relevance score, we initialize the relevance score as a learnable parameter. We refer to this method as \methodname{} w/o REL. We then compare the model's performance on commonsense reasoning and zero-shot tasks. As shown in \cref{table:update}, \methodname{} consistently outperforms \methodname{} w/o REL across all tasks, demonstrating the effectiveness of our proposed relevance module for dynamically adjusting sub-update contributions.

\section{Conclusion}

In this work, we introduce \methodname{}, a novel and versatile framework that enhances the performance of pretrained LLMs by dynamically adjusting the contributions of sub-updates based on their relevance to the input text or target output style.
Our experimental results across multiple language model backbones demonstrate the effectiveness of \methodname{} in improving model performance in both supervised fine-tuning and zero-shot settings while requiring far fewer trainable parameters. 
Moreover, \methodname{} also support controlled generation aligned with specific target style.
Importantly, our framework is architecture-agnostic and can be seamlessly integrated into any model containing a feed-forward layer.

\section*{Limitations}

While \methodname{} is designed to be compatible with any architecture that includes a feed-forward layer, our current study focuses solely on text-based LLMs, without evaluating its applicability to multimodal language model and image generation models.  
In future work, we plan to extend \methodname{} to a broader range of architectures, including multimodal and image generation models, to assess its versatility and impact across diverse architectures and domains.
Additionally, although our approach leverages sub-updates to refine model predictions, our work does not fully interpret the internal mechanisms of LLMs. Demystifying the internal states of these models remains a challenging task, and further research is needed.

\section*{Acknowledgements}
This research is supported by the award No. 2238940 from the Faculty Early Career Development Program (CAREER) of the National Science Foundation (NSF). The views and conclusions contained herein are those of the authors and should not be interpreted as necessarily representing the official policies, either expressed or implied, of the U.S. Government. The U.S. Government is authorized to reproduce and distribute reprints for governmental purposes notwithstanding any copyright annotation therein.

\bibliography{custom}

\appendix

\appendix



\section{Baselines}
\label{app:baselines}

We compare our approach against the following baselines: 
1) LoRA~\citep{hulora} leverages low-rank approximations of parameter matrices for parameter-efficient fine-tuning;
2) DExperts~\citep{liu2021dexperts} is a decoding-time method for controlled text generation that combines a pretrained language model with trained positive and negative label classifiers, using the difference in their classifiers' scores to adjust the model’s original logits;
3) GeDi~\citep{krause2021gedi} guides generation at each step by computing classification probabilities for all possible next tokens via Bayes rule;
4) PromptT5~\citep{raffel2020exploring} is a
pre-trained T5~\citep{raffel2020exploring} model optimized for prompt-based task solving. Following ~\citet{han2024word}, the model is prompted with``Complete this sentence so that it embodies a {positive/negative} sentiment:'' to generate sentiment-specific outputs;
5) LMSteer~\citep{han2024word} enables control over language model generation styles by applying a learnable linear transformation to output word embeddings.

\section{Implementation Details}
\label{app:hyper}

\begin{table}[!t]
  \centering
  \resizebox{\linewidth}{!}{%
  \begin{tabular}{l |  l c c  }
  \toprule
   &\textbf{Method}  & Learning Rate & Batch Size \\ \midrule
   \multirow{4}{0.6cm}{\rotatebox{90}{Qwen2.5-1.5B}} & LoRA (r=16) & 3e-4 & 16 \\
   & LoRA (r=32) & 3e-4 & 16\\
  &  \methodname{} (r=16) & 2e-3 & 16\\
  &  - \; w/o REL  & 3e-4 & 16\\
  &  \methodname{} (r=32) & 2e-3 & 16\\
  &  - \; w/o REL  & 3e-4 & 16 \\
   \midrule
   \multirow{4}{0.6cm}{\rotatebox{90}{Llama2-7B}} & LoRA (r=16) & 3e-4 & 16\\
  & LoRA (r=32) & 3e-4 & 16 \\
  & \methodname{}  (r=16) & 4e-2 & 16 \\
  &  - \; w/o REL  & 3e-4 & 16\\
  & \methodname{} (r=32) & 4e-2 & 16 \\
  &  - \; w/o REL  & 3e-4 & 16 \\
  \midrule
   \multirow{4}{0.6cm}{\rotatebox{90}{Llama3-8B}} & LoRA (r=16) &   3e-4 & 16  \\
    & LoRA (r=32) &   3e-4 & 16   \\
    & \methodname{}  (r=16)  & 4e-2 & 16\\
  & \methodname{} (r=32) & 4e-2 & 16\\
  \bottomrule
  \end{tabular}
  }
  \caption{Final hyperparameter settings chosen after tuning over the hyperparameter search spaces.}
  \label{table:hyperparam}
  \end{table}

For all experiments, we employ the AdamW optimizer with $\beta_1=0.9$ and $\beta_2=0.999$. We use a cosine learning rate schedule, with a warmup ratio of $0.2$, and decay the final learning rate down to $20\%$ of the peak learning rate. 
Due to computing resource constraints, we limited our hyperparameter tuning to the learning rate and batch size. Specifically, the search space for the learning rate is \{4e-2, 2e-3, 3e-4, 2e-5\}, and for the batch size, it is \{16, 32\}. \cref{table:hyperparam} lists the final hyperparameters chosen.
Following \citet{hu2023llm}, we set the number of epochs to 3 for the supervised commonsense reasoning tasks, and we use 6 epochs for the zero-shot setting, considering the smaller size of the instruction-tuning dataset. All experiments are conducted on A100 80G GPUs.
\begin{table}[!ht]
  \centering
  \resizebox{\linewidth}{!}{%
  \begin{tabular}{ c | p{0.9\linewidth} }
  \toprule
    Steering Value  & Generation \\
   \midrule
    \multirow{2}{*}{0} & \underline{According to other witnesses the men}  and women were killed while being questioned. \\ \hline
    \multirow{2}{*}{-3} & \underline{According to other witnesses the men}  were taken to a nearby village to be questioned. \\ \hline
    \multirow{2}{*}{-5} & \underline{According to other witnesses the men} and women were escorted into the home. \\
  \bottomrule
  \end{tabular}
  }
  \caption{Qualitative examples of toxic language suppression using \methodname{}. \underline{Underlined text} indicates the input provided to the model, while the remaining text is generated by \methodname{}.}
  \label{table:toxic_ex}
  \end{table}
\begin{table*}[b!]
  \centering
  \resizebox{\linewidth}{!}{%
  \begin{tabular}{p{\linewidth}}
    \toprule
    \textbf{Positive Sentiment} \\ \midrule
   "Amazing",
    "Astonishing",
    "Breathtaking",
    "Brilliant",
    "Delightful",
    "Exceptional",
    "Extraordinary",
    "Fantastic",
    "Glorious",
    "Incredible",
    "Magnificent",
    "Marvelous",
    "Outstanding",
    "Phenomenal",
    "Remarkable",
    "Spectacular",
    "Splendid",
    "Stupendous",
    "Superb",
    "Terrific",
    "Thrilling",
    "Unbelievable",
    "Wonderful",
    "Bliss",
    "Brilliance",
    "Euphoria",
    "Excellence",
    "Genius",
    "Glory",
    "Harmony",
    "Joy",
    "Miracle",
    "Perfection",
    "Radiance",
    "Splendor",
    "Triumph",
    "Wonder",
    "Adore",
    "Amaze",
    "Astonish",
    "Celebrate",
    "Elevate",
    "Enchant",
    "Energize",
    "Excite",
    "Inspire",
    "Marvel",
    "Overjoy",
    "Rejoice",
    "Thrill",
    "Absolutely",
    "Astonishingly",
    "Exceptionally",
    "Exquisitely",
    "Fantastically",
    "Incredibly",
    "Marvelously",
    "Remarkably",
    "Spectacularly",
    "Stunningly",
    "Unbelievably",
    "Wonderfully"
    \\ \midrule
    \textbf{Negative Sentiment} \\ \midrule
   "abysmal",
    "appalling",
    "atrocious",
    "awful",
    "catastrophic",
    "cursed",
    "damning",
    "dismal",
    "disastrous",
    "disturbing",
    "evil",
    "ghastly",
    "horrendous",
    "horrible",
    "horrific",
    "hopeless",
    "loathsome",
    "lousy",
    "monstrous",
    "nauseating",
    "nightmarish",
    "outrageous",
    "pathetic",
    "pitiful",
    "revolting",
    "shameful",
    "shocking",
    "terrible",
    "terrifying",
    "toxic",
    "tragic",
    "unbearable",
    "unforgivable",
    "unlivable",
    "unthinkable",
    "unworthy",
    "vicious",
    "wretched",
    "abominable",
    "deplorable",
    "disgusting",
    "grotesque",
    "hideous",
    "inexcusable",
    "insufferable",
    "intolerable",
    "malevolent",
    "nefarious",
    "offensive",
    "repugnant",
    "scandalous",
    "sinister",
    "vile",
    "villainous",
    "wicked",
    "fearful",
    "bitter",
    "frustrated",
    "miserable",
    "lonely",
    "selfish"
    \\ \midrule
    \textbf{Toxic Words} \\ \midrule
    "fuck", "shit", "hell", "abbo", "abo", "beeyotch", "biatch", "bitch", "chinaman", "chinamen", "chink", "coolie", "coon", "crazie", "crazy", "crip", "cuck", "cunt", "dago", "daygo", "dego", "dick", "douchebag", "dumb", "dyke", "eskimo", "fag", "faggot", "fatass", "fatso", "gash", "gimp", "gip", "golliwog", "gook", "gyp", "gypsy", "half-breed", "halfbreed", "heeb", "homo", "hooker", "idiot", "insane", "insanitie", "insanity", "jap", "kaffer", "kaffir", "kaffir", "kaffre", "kafir", "kike", "kraut", "lame", "lardass", "lesbo", "lunatic", "mick", "negress", "negro", "nig", "nig-nog", "nigga", "nigger", "nigguh", "nip", "pajeet", "paki", "pickaninnie", "pickaninny", "prostitute", "pussie", "pussy", "raghead", "retard", "sambo", "shemale", "skank", "slut", "soyboy", "spade", "sperg", "spic", "spook", "squaw", "street-shitter", "tard", "tits", "titt", "trannie", "tranny", "twat", "wetback", "whore", "wigger", "wop", "yid", "zog"
     \\
    \bottomrule
  \end{tabular}}
  \caption{The list of attribute tokens employed for conditional generation. Please note that the words listed under "toxic words" could be offensive to read.}
  \label{tab:attributes}
\end{table*}

\section{Conditional Generation}
\label{app:cond_gen}

\subsection{Evaluation Metrics}
\label{app:evaluation}

We follow the evaluation protocol in \citet{han2024word} and assess the quality of the generated text using fluency (perplexity) and diversity metrics. Specifically, fluency is measured by the perplexity score computed with GPT2-XL, while diversity is quantified using the proportion of distinct \{1, 2, 3\}-grams for Dist-\{1, 2, 3\}.

For sentiment evaluation, we employ HuggingFace’s sentiment classifier~\citep{wolf2020transformers} and compute the average percentage of positive outputs per prompt, which we refer to as the ``Positivity'' score.

For toxic language suppression, we evaluate the toxicity of the generated outputs using the Perspective API~\footnote{https://perspectiveapi.com/} to grade the text.
We consider two metrics: the maximal toxicity of generations on each prompt averaged across prompts (``Max. toxicity'') and the
averaged probability that a generation exceeds a toxicity threshold of $0.5$ (``Toxicity prob.'').

\subsection{Construction of StyleDataset}
\label{app:style}

\begin{table*}[b!]
  \centering
  \label{tab:attributes}
  \resizebox{\linewidth}{!}{%
  \begin{tabular}{p{\linewidth}}
    \toprule
    You are a highly intelligent and detail-oriented assistant, renowned for your ability to leverage extensive knowledge and analyze questions carefully. You think through each problem step by step and always provide structured, clear, and concise answers.\\\\

Given a sentence, your task is to determine its most prominent stylistic property and then generate a short list of descriptive attribute words that capture that style. The attributes may relate to properties such as formality, technicality, narrative style, sentiment, humor, etc.\\\\

Below are some examples:\\\\

Example 1 – Formality/Professional Tone:\\
Sentence:\\
"Dear Sir or Madam, please be advised that your account will be debited on the next business day."\\
Attribute words:\\
"formal, professional, courteous, business-like"\\\\

Example 2 – Informal/Conversational Tone:\\
Sentence:\\
"Hey, what's up? Wanna hang out later?"\\
Attribute words:\\
"informal, casual, friendly, colloquial"\\\\

Example 3 – Technical/Scientific Language:\\
Sentence:\\
"The quantum state of the particle was determined via the Fourier transform of its wavefunction."\\
Attribute words:\\
"technical, scientific, precise, academic"\\\\

Example 4 – Narrative/Story-like Style:\\
Sentence:\\
"Once upon a time, in a land shrouded in mystery, a young adventurer embarked on a quest."\\
Attribute words:\\
"narrative, descriptive, imaginative, literary"\\\\

Example 5 – Humorous/Witty Tone:\\
Sentence:\\
"I tried to catch some fog yesterday, but I mist."\\
Attribute words:\\
"humorous, playful, pun-based, witty"\\\\

Now, for the following sentence, provide a list of attribute words that best describe its style. You should answer with the following structure:\\\\

Explain: \\
<explanation>\\
Attribute words: \\
<answer>\\\\

Sentence:"\{sentence\}"\\
Explain: \\
    \bottomrule
  \end{tabular}}
  \caption{The instruction for prompting Llama3.1-8B-instruct~\citep{dubey2024llama} for generating list of attribute words for each sentence.}
  \label{tab:style_data}
\end{table*}

\begin{table*}[!t]
  \centering
  \resizebox{\linewidth}{!}{%
  \begin{tabular}{p{0.7\linewidth} p{0.3\linewidth}}
  \toprule
   \textbf{Sentence} & \textbf{Style}   \\ \midrule
   On behalf of the team, I am pleased to announce that Spring Boot 1.5.4 has been released and is available now from repo.spring.io and Maven Central. & formal, technical, professional, announcement-like, dry, informative \\ \hline
   Step 1: Preheat oven to 400F. & technical, instructional, straightforward, directive \\ \hline
   I like to hear music while I cook, but nothing too headbangy any more. Curtis Mayfield, Marvin Gaye, pre-disco funk, Isaac Hayes and Brothers Johnson and I'm happy. & casual, conversational, personal, emotive \\ \hline
   You could follow Alt News posts either via our Facebook page or by following us on Twitter or by subscribing to our E-mail updates. & direct, informative, instructive, utilitarian \\ \hline
   It seems not everything is more expensive these days.&oberving, casual, sarcastic, conversational, observational, casual.\\ \hline
   This photograph, taken by an astronaut aboard the International Space Station, shows the straight line of the Corinth Canal as it crosses a narrow isthmus between mainland Greece (right) and the Peloponnese Peninsula. The canal cuts through the narrowest part of the isthmus of Corinth. The towns of Corinth and Isthmia stand near the west and east ends (north is to the upper right). Near the center of the image, a highway crosses the canal and connects Athens to the Peloponnese.& informative, formal, descriptive, geographic \\ \hline
   The Liberal position is consistent with Bill C-60, their 2005 copyright bill that linked the digital lock rules to actual copyright infringement and did not establish a ban on the tools that can be used to circumvent digital locks.&formal, analytical, technical, informative, explanatory, objective\\
  \bottomrule
  \end{tabular}}
  \caption{Example data instances from \textit{StyleDataset}.}
  \label{tab:ex_style_data}
  \end{table*}

\cref{tab:style_data} shows the instruction used to prompt the Llama3.1-8B-instruct~\citep{dubey2024llama} model to generate a short list of descriptive attributes capturing a sentence's style. We also provide several example instances from \textit{StyleDataset} in \cref{tab:ex_style_data}.

\section{Qualitative Results}
\label{app:qualitative}

\cref{table:toxic_ex} illustrates how varying the steering value $s$ reduce the toxicity of generated text.


\end{document}